\documentclass{article}
\usepackage[square,numbers,sort&compress]{natbib}
\usepackage{amsmath,graphicx,amssymb,natbib,enumitem,xcolor,booktabs,threeparttable,multirow,adjustbox}
\usepackage[preprint]{spconf}
\usepackage{booktabs, multirow, threeparttable}

\usepackage[colorlinks=true, urlcolor=blue, pdfborder={0 0 0}]{hyperref}

\title{Cross-Lingual Interleaving for Speech Language Models}
\name{Adel Moumen\thanks{This work was supported by the Cambridge Commonwealth, European \& International Trust. }, Guangzhi Sun, Philip C. Woodland}
\address{Department of Engineering, University of Cambridge, UK \\ \texttt{\{am3303,gs534,pcw117\}@cam.ac.uk}}

\begin{document}
\ninept
\maketitle

\begin{abstract}

Spoken Language Models (SLMs) aim to learn linguistic competence directly from speech using discrete units, widening access to Natural Language Processing (NLP) technologies for languages with limited written resources. However, progress has been largely English\mbox{-}centric due to scarce spoken evaluation benchmarks and training data, making cross\mbox{-}lingual learning difficult. We present a cross\mbox{-}lingual interleaving method that mixes speech tokens across languages without textual supervision. We also release an EN\mbox{--}FR training dataset, \emph{TinyStories} (\(\sim\!42\)k hours), together with EN\mbox{--}FR spoken \emph{StoryCloze} and \emph{TopicCloze} benchmarks for cross\mbox{-}lingual semantic evaluation, both synthetically generated using GPT\mbox{-}4. On $360$M and $1$B SLMs under matched training\mbox{-}token budgets, interleaving improves monolingual semantic accuracy, enables robust cross\mbox{-}lingual continuation, and strengthens cross\mbox{-}lingual hidden\mbox{-}state alignment. Taken together, these results indicate that cross\mbox{-}lingual interleaving is a simple, scalable route to building multilingual SLMs that understand and converse across languages. All resources are made open-source to support reproducibility\footnote{HuggingFace Datasets Available at: \href{https://huggingface.co/datasets/Adel-Moumen/Spoken_TinyStories}{Adel-Moumen/Spoken\_TinyStories}}.
\end{abstract}

\begin{keywords}
Spoken language models, cross\mbox{-}lingual learning, interleaving, multilinguality
\end{keywords}

\section{Introduction}
\label{sec:intro}

Spoken Language Models (SLMs) aim to learn language directly from speech, without relying on textual supervision. This line of work emerged from \emph{`textless' NLP} \cite{lakhotia2021generative}, which combines discrete speech representations \cite{mousavi2025discrete} and neural language modelling \cite{brown2020language,Radford2018ImprovingLU} to capture both acoustic and linguistic regularities from audio alone. Within this framework, \emph{Generative Spoken Language Modelling} (GSLM) formalises the objective by training an autoregressive LM on sequences of discrete speech units and reconstructing waveforms with a unit vocoder, with the motivation of broadening access to NLP in settings where written resources are scarce.

Despite rapid progress, the \emph{cross-lingual} setting, which aims to train a single SLM over multiple languages, remains comparatively under-explored~\citep{nguyen2024spiritlm,zeng2024scalingspeechtextpretrainingsynthetic,futami2025scheduled}. Two practical bottlenecks persist: (i) evaluations are predominantly English-centric, hindering measurement of cross-lingual semantic competence; and (ii) training methods that encourage sharing across languages often depend on \emph{text}, for example via speech–text \emph{interleaving} in Text–Speech LMs (TSLMs)~\citep{chou-etal-2023-toward,nguyen2024spiritlm,cuervo2025textspeechlanguagemodelsimproved,zeng2024scaling}, which is not directly compatible with \emph{`textless'} SLMs. As a result, the original promise of \emph{`textless'} NLP learning from audio alone at scale and across languages has not yet been fully realised. 

We address these limitations by introducing a \emph{cross-lingual interleaving} scheme that mixes speech tokens from different languages within the same training sequence, using sentence-level alignments but no text tokens. Exposure to interleaved multilingual contexts encourages a shared representational subspace and promotes transfer, while remaining compatible with pure \emph{`textless'} pipelines. To make this practical and measurable, we curate bilingual, sentence-aligned spoken resources and new spoken semantic benchmarks. Our contributions are as follows:
\begin{enumerate}[label=(\roman*)]
    \item We propose a cross\mbox{-}lingual interleaving strategy that alternates speech segments from different languages within a single sequence, promoting a shared representational subspace.
    \item We release a French–English, sentence-aligned spoken corpus derived from \emph{TinyStories}~\citep{eldan2023tinystories} (approximately $42$k hours in total) together with bilingual spoken \emph{StoryCloze} and \emph{TopicCloze} benchmarks~\citep{mostafazadeh-etal-2016-corpus} for semantic evaluation synthetically generated using GPT-4~\citep{achiam2023gpt}.
    \item We show on $360$M and $1$B parameter SLMs that the method yields \emph{positive transfer} to monolingual semantic tasks and strong \emph{cross-lingual} performance, under a matched training-token budget. Additionally, representation analyses indicate stronger cross-lingual alignment in the hidden states.
\end{enumerate}

\noindent Section~\ref{sec:related_work} situates our approach with respect to SLMs, speech–text interleaving, and multilingual training. Section~\ref{sec:spoken_language_modelling} formalises the modelling setup and the cross-lingual interleaving scheme; Sections~\ref{sec:dataset}–\ref{sec:experimental_results} present datasets, experiments, and analyses.

\section{Related Work}
\label{sec:related_work}

\subsection{Speech Language Models} 

Early SLMs instantiate the \emph{`textless'} pipeline by modelling discrete speech units and decoding back to waveforms. GSLM~\citep{lakhotia2021generative} trains a decoder\nobreakdash-only LM over units and synthesises speech with a unit\nobreakdash-based vocoder; pGSLM~\cite{kharitonov2021text} augments units with prosodic tokens to improve expressiveness; and dGSLM~\citep{nguyen2023generative} introduces a dual\nobreakdash-tower variant for two\nobreakdash-channel dialogues. While these systems can produce locally fluent speech, long\nobreakdash-form coherence is challenging. SpeechSSM~\citep{park2024long} explores state\nobreakdash-space models~\citep{gu2021efficiently} for long\nobreakdash-form generation, TWIST~\citep{hassid2023textually} shows that initialising from a pre\nobreakdash-trained text Large Language Model (LLM) can boost semantics, and AudioLM~\citep{borsos2023audiolm} cascades semantic and acoustic token streams. More recently, Slamming~\citep{maimon2025slammingtrainingspeechlanguage} provides a competitive compute\nobreakdash-efficient recipe, and Align\nobreakdash-SLM~\citep{lin2024align} fine\nobreakdash-tunes SLMs with direct preference optimisation. Our work stays within the speech\nobreakdash-only paradigm but targets \emph{cross\nobreakdash-lingual} sharing through a new learning method.

\subsection{Joint Speech–Text Interleaving} 

TSLMs extend SLMs with text to encourage cross\nobreakdash-modal transfer. VoxtLM/SUTLM~\citep{chou-etal-2023-toward} jointly trains on ASR, TTS, and continuation, and identifies \emph{interleaving} (switching between text and speech inside a sequence) as particularly effective for building a shared space. Spirit-LM~\citep{nguyen2024spiritlm} scales this recipe and emphasises interleaving as a key quality driver; SmolTolk~\citep{cuervo2025textspeechlanguagemodelsimproved} proposes architecture refinements for vocabulary expansion; and GLM4-Voice~\citep{zeng2024scalingspeechtextpretrainingsynthetic} leverages a text\nobreakdash-to\nobreakdash-token model~\citep{zeng2024scaling}  to generate audio tokens  on\nobreakdash-the\nobreakdash-fly, sidestepping explicit aligners~\citep{radford2022robustspeechrecognitionlargescale,pratap2024scaling}. However, these approaches depend on text and interleave within a single language, limiting direct applicability to textless, cross\nobreakdash-lingual SLM training. We instead interleave \emph{speech tokens across languages} without introducing text.

\subsection{Cross-Lingual Training}  

Multilingual training for speech LMs is gaining traction but remains under\nobreakdash-evaluated on spoken semantic understanding beyond English. Spirit-LM~\citep{nguyen2024spiritlm} includes multiple languages at scale but reports only English spoken evaluations; GLM4-Voice~\citep{zeng2024scalingspeechtextpretrainingsynthetic} trains on Chinese and English yet evaluates English semantics; and scheduled interleaving for speech\nobreakdash-to\nobreakdash-speech translation~\citep{futami2025scheduled} interleaves \emph{monolingually} with no cross\nobreakdash-language interaction during continuation. In text, mixing languages can induce positive transfer~\citep{yoo-etal-2025-code-switching}. Our method brings this intuition to the \emph{`textless'} speech regime by interleaving sentence\nobreakdash-aligned speech sequences across languages, enabling both cross\nobreakdash-lingual abilities and improved monolingual semantics under a matched training\nobreakdash-token budget.

\section{Spoken Language Modelling}
\label{sec:spoken_language_modelling}

The task of modelling spoken language from raw audio typically comprises three stages: (i) quantising a waveform $\mathbf{x}\in\mathbb{R}^{T}$ into a sequence of $L$ discrete speech units with $L\ll T$; (ii) training an SLM under a next\mbox{-}token objective on the quantised sequence; and (iii) synthesising a waveform from the discrete units via a neural decoder. We first describe speech tokenisation, then the SLM objective, and finally our cross\mbox{-}lingual interleaving scheme.

\subsection{Speech Language Models}
\label{subsec:slm}

\noindent\textbf{Speech tokenisers.}
Speech tokenisers transform raw waveforms into discrete representations.
Let $\mathcal{X}$ denote the domain of audio samples and let a waveform be $\mathbf{x}=(x_{1},\dots,x_{T})\in\mathcal{X}^{T}$.
An encoder $f:\mathbb{R}^{T}\!\rightarrow\!\mathbb{R}^{T'\times d_e}$ produces a sequence of continuous features $f(\mathbf{x})=(\mathbf{v}_{1},\dots,\mathbf{v}_{T'})$, sampled at a lower frame rate $T'\ll T$.
A quantiser $q$ maps these to discrete units $\,\mathbf{s}=(s_{1},\dots,s_{T'})$ with $s_i\in\{1,\dots,K\}$, where $K$ is the codebook size.
Common choices include $k$\mbox{-}means and residual vector quantisation (RVQ).

In this work we adopt Mimi~\citep{defossez2024moshispeechtextfoundationmodel}, a neural audio codec~\citep{zeghidour2021soundstream,defossez2022high} implemented as a convolutional auto\mbox{-}encoder with RVQ in the latent space with semantic distillation from the WavLM representations~\citep{chen2022wavlm}.
Following prior evidence that early codebooks capture semantics, we \emph{only} model the first (semantic) codebook during language modelling. 

\medskip
\noindent\textbf{Speech language models.}
Given a sequence of $L$ speech tokens $\mathbf{s}=(s_{1},\dots,s_{L})$ over vocabulary $\mathcal{V}_{s}$, an SLM parameterises
\begin{equation}
p_{\theta}(\mathbf{s})=\prod_{i=1}^{L} p_{\theta}\!\left(s_{i}\mid s_{<i}\right).
\end{equation}
Decoder\mbox{-}only transformers~\citep{transformer} are typically used as SLM and trained by minimising the autoregressive negative log\mbox{-}likelihood
\begin{equation}
\mathcal{L}_{\mathrm{LM}}=-\sum_{i=1}^{L}\log p_{\theta}\!\left(s_{i}\mid s_{<i}\right).
\label{eq:lm_loss}
\end{equation}
Each token $s_i$ is embedded via $E\in\mathbb{R}^{|\mathcal{V}_{s}|\times d}$ to $\mathbf{e}_{i}=E[s_{i}]$.
A stack of $m$ causal transformer blocks maps $(\mathbf{e}_{1},\dots,\mathbf{e}_{L})$ to contextual states $(\mathbf{h}_{1},\dots,\mathbf{h}_{L})$, with $\mathbf{h}_{i}$ depending only on $s_{<i}$.
A projection $U\in\mathbb{R}^{d\times|\mathcal{V}_{s}|}$ yields logits $\boldsymbol{\ell}_{i}=U^{\top}\mathbf{h}_{i}$ and the predictive distribution $p_{\theta}(s_{i}\!\mid\! s_{<i})=\mathrm{softmax}(\boldsymbol{\ell}_{i})$.

\subsection{Cross-Lingual Interleaving Training}
\label{subsec:interleaving}

Standard SLM training predicts the next speech token within a monolingual sequence.
Interleaving has been proposed in TSLMs to bridge text--speech modality gaps by switching at word boundaries~\citep{chou-etal-2023-toward,nguyen2024spiritlm,zeng2024scalingspeechtextpretrainingsynthetic,cuervo-marxer-2024-scaling}, but such methods rely on text tokens and are thus incompatible with \emph{textless} SLMs; moreover, they do not readily support mixing multiple languages~\citep{yoo-etal-2025-code-switching}.

We formulate a \emph{cross\mbox{-}lingual interleaving} scheme that operates directly on speech tokens by interleaving segments from multiple languages.
Assuming sentence\mbox{-}level segmentation and cross\mbox{-}lingual sentence alignment, let $A^{(l)}_{j}$ denote the $j$th segment in language $l$, represented as a sequence of speech tokens.
Given a set of languages $\mathcal{L}$, we sample a language label $l_j \in \mathcal{L}$ for each segment index $j$ and define the interleaved sequence as
\begin{equation}
s = \big(A^{(l_1)}_{1}\,\Vert\,A^{(l_2)}_{2}\,\Vert\,A^{(l_3)}_{3}\,\Vert\,\dots\big),
\label{eq:interleaving_eq}
\end{equation}
where $\Vert$ denotes concatenation at sentence boundaries.\footnote{Word\mbox{-}level switching is also possible when word\mbox{-}level alignments are available; unless stated otherwise, we use sentence\mbox{-}level switching.}
We then train with the standard next\mbox{-}token prediction objective on $s$, so that each prediction is conditioned on preceding segments that may come from different languages.
Since Eq.~\ref{eq:interleaving_eq} is language-agnostic, it can in principle be applied to any language for which speech tokens are available; this is consistent with observations that code-switching-style training can transfer beyond distant languages~\citep{yoo-etal-2025-code-switching}.
Having defined the mechanism, we next describe the sentence\mbox{-}aligned spoken corpora that make such training feasible (see Section~\ref{sec:dataset}).

\section{Cross-Lingual Spoken Datasets}
\label{sec:dataset}
\begin{table}[t]
\vspace{-6pt}
\centering
\footnotesize
\setlength{\tabcolsep}{3pt}
\renewcommand{\arraystretch}{0.95}
\caption{Cross\mbox{-}Lingual \emph{TinyStories} per\mbox{-}language statistics. Durations are computed over the audio stories in each language.}
\label{tab:csts_lang_stats}
\begin{adjustbox}{max width=\columnwidth}
\begin{tabular}{llrrrr}
\toprule
\textbf{Split} & \textbf{Lang} & \textbf{Total (h)} & \textbf{Avg (s)} & \textbf{Min (s)} & \textbf{Max (s)} \\
\midrule
Train       & EN & 20294.90 & 55.52 &  3.12 & 140.48 \\
            & FR & 21120.66 & 57.77 &  3.12 & 135.84 \\
Validation  & EN &   298.04 & 55.22 &  4.64 & 121.44 \\
            & FR &   311.64 & 57.74 &  4.08 & 120.64 \\
\bottomrule
\end{tabular}
\end{adjustbox}
\end{table}

\begin{table}[t]
\vspace{-6pt}
\centering
\footnotesize
\setlength{\tabcolsep}{3pt}
\renewcommand{\arraystretch}{0.95}
\caption{Cross\mbox{-}Lingual spoken \emph{StoryCloze} and spoken \emph{TopicCloze} per\mbox{-}language statistics. Durations are computed over the audio stories in each language.}
\label{tab:ssc_lang_stats}
\begin{adjustbox}{max width=\columnwidth}
\begin{tabular}{llrrrr}
\toprule
\textbf{Split} & \textbf{Lang} & \textbf{Total (h)} & \textbf{Avg (s)} & \textbf{Min (s)} & \textbf{Max (s)} \\
\midrule
StoryCloze & EN & 15.68 & 15.09 & 6.32 & 24.96 \\
           & FR & 16.75 & 16.12 & 5.12 & 26.16 \\
TopicCloze & EN & 15.64 & 15.05 & 5.92 & 24.08 \\
           & FR & 16.76 & 16.13 & 6.48 & 27.20 \\
\bottomrule
\end{tabular}
\end{adjustbox}
\end{table}

Training SLMs is compute\mbox{-}intensive and data\mbox{-}hungry: effective systems require large speech corpora, while cross\mbox{-}lingual interleaving (see Section~\ref{subsec:interleaving}) additionally demands alignment.
To make such training practical, we release a French–English, sentence\mbox{-}aligned spoken corpus derived from \emph{TinyStories}~\citep{eldan2023tinystories}, totalling approximately $42$k hours of speech (see Table~\ref{tab:csts_lang_stats} for statistics).
In addition, we introduce bilingual spoken versions of \emph{StoryCloze} and \emph{TopicCloze} for evaluating cross\mbox{-}lingual semantic understanding (Table~\ref{tab:ssc_lang_stats}). We next summarise how these datasets are constructed.

\subsection{Cross-Lingual TinyStories}
\label{subsec:cross_lingual_tinystories}
\emph{TinyStories} is a synthetic corpus of short narratives whose vocabulary targets 3\mbox{–}4\mbox{-}year\mbox{-}olds, generated by large language models to probe commonsense reasoning in text LMs.
Its spoken variant has proved effective for SLM training, improving both modelling and generative metrics~\citep{cuervo-marxer-2024-scaling,maimon2025slammingtrainingspeechlanguage}.
Stories are self\mbox{-}contained, causally structured, and fit within a typical SLM context windows.

In order to preserve cross\mbox{-}lingual semantic coherence, we construct sentence\mbox{-}level alignments that preserve order and meaning.
A high\mbox{-}quality MT system (GPT\mbox{-}4)~\citep{achiam2023gpt} translates each sentence with full story context; the entire validation set and $600$k training sentences are translated into French while maintaining English alignment.
We then synthesise speech with a multi\mbox{-}speaker TTS system~\citep{zeghidour2025streamingsequencetosequencelearningdelayed}\footnote{https://github.com/kyutai-labs/delayed-streams-modeling} (a $1.6$B model using delayed\mbox{-}streams modelling first introduced in Moshi~\citep{defossez2024moshispeechtextfoundationmodel}).

We subsample $500$ cross\mbox{-}lingual stories from the EN-FR \emph{sTinyStories} validation set, for each candidate voice we synthesise EN–FR pairs with the same voice.
Using cosine similarity between Speaker Verification (SV)~\citep{chen2022wavlm}\footnote{HuggingFace Model: microsoft/wavlm-base-sv} embeddings, we retain voices scoring $>0.90$ on average across paired sentences, yielding $44$ voices (26 female, 18 male).
We synthesise the final sets with these $44$ voices, reusing the same speaker across languages  to promote cross\mbox{-}lingual speaker consistency.
The \emph{sTinyStories} corpus statistics are given in Table~\ref{tab:csts_lang_stats}.

\subsection{Cross-Lingual StoryCloze and TopicCloze}
\label{subsec:cross_lingual_storycloze_topiccloze}
Assessing cross\mbox{-}lingual semantic abilities of SLMs is challenging due to the scarcity of spoken benchmarks beyond English~\citep{nguyen2024spiritlm,zeng2024scalingspeechtextpretrainingsynthetic}.
Inspired by \emph{StoryCloze}~\citep{mostafazadeh-etal-2016-corpus} and its spoken instantiation~\citep{hassid2023textually}, we introduce bilingual, sentence\mbox{-}aligned spoken variants—\emph{sSC} and \emph{sTC}—in French and English.

Each benchmark comprises $4$k five\mbox{-}sentence stories: the model observes the first four sentences and must assign higher probability (log\mbox{-}likelihood) to the true ending than to an adversarial alternative.
\emph{sSC} uses semantically incompatible endings, probing causal/temporal commonsense; \emph{sTC} draws negatives from different topics, probing topical coherence.

Corpus construction mirrors that in Section~\ref{subsec:cross_lingual_tinystories}: sentences are translated with GPT\mbox{-}4 using full story context, then synthesised with the same TTS system to preserve timing and prosody. To minimise confusion from speaker variability, we use a single female speaker (the top SV performer on validation) across both languages. Dataset statistics are given in Table~\ref{tab:ssc_lang_stats}.

\section{Experimental Setup}
\label{sec:experimental_setup}

Having introduced the interleaving mechanism and the datasets that support it, we now describe our training, and evaluation protocol.

\subsection{Models and training}

We use the Mimi tokeniser~\citep{defossez2024moshispeechtextfoundationmodel} at $12.5$~Hz with vocabulary size $K{=}2048$ and RVQ with $32$ codebooks; only the first (semantic) codebook is modelled.
SLMs are initialised from a $1$B\mbox{-}parameter Llama~3.2 family checkpoint~\citep{dubey2024llama} and a $360$M\mbox{-}parameter Qwen2 family checkpoint~\citep{team2024qwen2}, following TWIST~\citep{hassid2023textually}, with a $2048$\mbox{-}token context window ($\approx 2.73$ minutes of speech). These initialisations are used for the $1$B and $360$M models. The textual embedding tables are replaced with audio embeddings for the new tokens.

Training uses $4$ H$100$ ($80$~GiB) GPUs, per\mbox{-}GPU batch $153{,}600$ tokens (total $614{,}400$ tokens/step).
The Adam~\citep{kingma2014adam} optimiser is used with settings of $(\beta_{1},\beta_{2})=(0.9,0.98)$, gradient clipping $1.0$, weight decay $0.1$, and a linear warm\mbox{-}up of $5\%$ to $5{\times}10^{-4}$ followed by linear decay.
Inputs are packed by concatenating samples until the target length is reached.

Training proceeds in three stages to isolate the effect of interleaving: \textit{(1)} EN\mbox{-}only pre\mbox{-}training for $50$k steps to simulate a high\mbox{-}resource SLM; \textit{(2)} cross\mbox{-}lingual interleaving for $20$k steps with a language sampling probability of $0.5$ to encourage positive transfer from the high\mbox{-}resource language (EN) to a new language (FR); and \textit{(3)} alternating monolingual fine\mbox{-}tuning in FR and EN for $15$k steps to enable monolingual generation in both languages.
Unless noted, the $1$B and $360$M models follow the same schedule.
All comparisons are made with a matched training\mbox{-}token budget, meaning that each system sees the same number of speech tokens (and hence the same total speech duration at $12.5$~Hz).
By matching the total training\mbox{-}token budget, we ensure that any gains arise from interleaving rather than from data volume or sampling differences.

\subsection{Baselines}
We compare to: \textbf{(i) FR\mbox{-}only monolingual}, trained solely on French; and \textbf{(ii) EN{+}FR without interleaving}, trained on mixed data but restricted to monolingual sequences. Both baselines follow the same protocol as mentioned before.

\subsection{Datasets}

For EN we use \emph{LibriHeavy}~\citep{kang2024libriheavy} ($56$k hours) plus our EN \emph{sTinyStories} (see Section~\ref{sec:dataset}) split for a total of $76$k hours.
For FR we use our FR \emph{sTinyStories} ($\sim 21$k hours) split.
Interleaving uses the full \emph{sTinyStories} sentence\mbox{-}aligned EN–FR corpus described above.

\subsection{Evaluation}

\noindent \textbf{\emph{sBLiMP}}~\citep{nguyen2020zero}
tests syntactic competence: for each grammatical/ungrammatical pair, the model should assign higher probability to the grammatical item.

\noindent \textbf{\emph{sWUGGY}}~\citep{nguyen2020zero}
measures lexical discrimination by preferring a real word over a phonotactically similar nonce word.

\noindent \textbf{\emph{sSC} and \emph{sTC}}
(see Section~\ref{sec:dataset}) evaluate semantic understanding by requiring preference for the true ending over an adversarial alternative.

\noindent \textbf{Cross\mbox{-}lingual evaluation.}
We evaluate \emph{cross\mbox{-}lingual} \emph{sSC} and \emph{sTC} (EN$\!\rightarrow\!$FR, FR$\!\rightarrow\!$EN): prompts are in one language and continuations in the other. A robust cross\mbox{-}lingual SLM should achieve comparable performance across these conditions.

\section{Experimental Results}
\label{sec:experimental_results}

The results are organised around three questions: (i) does interleaving yield \emph{positive transfer} to monolingual performance? (ii) does it induce \emph{cross\mbox{-}lingual abilities}? and (iii) does it help build a shared structure of cross\mbox{-}lingual speech representations? The findings are reported for both $360$M and $1$B SLMs. 

\begin{table*}[t]
\vspace{-6pt}
\centering
\setlength{\tabcolsep}{4pt}
\caption{Cross\mbox{-}lingual results (accuracy, \%). Best within each sub\mbox{-}block in \textbf{bold}, second\mbox{-}best \underline{underlined}. Interleaving shows cross-lingual transfer in monolingual generation and cross-lingual performances while preserving semantic, syntactic, and lexical SLMs performances. The EN+FR baseline displays limited cross-lingual abilities, despite showing positive transfer.}
\label{tab:crosslingual_results}
\resizebox{1.0\textwidth}{!}{%
\begin{tabular}{lccccccccccccc}
\toprule
\multicolumn{1}{l}{Training setup} & \# Parameters & \multicolumn{2}{c}{Token budget} & \multicolumn{4}{c}{\emph{StoryCloze (sSC)}} & \multicolumn{4}{c}{\emph{TopicCloze (sTC)}} & \multicolumn{2}{c}{\emph{ZeroSpeech}} \\
\cmidrule(lr){3-4}\cmidrule(lr){5-8}\cmidrule(lr){9-12}\cmidrule(lr){13-14}
 &  & EN & FR & EN & FR & EN$\to$FR & FR$\to$EN & EN & FR & EN$\to$FR & FR$\to$EN & sBLiMP (EN) & sWUGGY (EN) \\
\midrule
random & -- & -- & -- & 50.00 & 50.00 & 50.00 & 50.00 & 50.00 & 50.00 & 50.00 & 50.00 & 50.00 & 50.00 \\
\midrule
Baseline EN & 1\text{B} & 46.08 & -- & \underline{56.06} & -- & -- & -- & 66.43 & -- & -- & -- & \underline{61.96} & \textbf{69.92} \\
Baseline EN & 360\text{M} & 46.08 & -- & \textbf{56.38} & -- & -- & -- & 65.20 & -- & -- & -- & 61.62 & \underline{69.32} \\
\midrule
Baseline FR & 1\text{B} & -- & 15.36 & -- & 55.31 & -- & -- & -- & 67.07 & -- & -- & -- & -- \\
Baseline FR & 360\text{M} & -- & 15.36 & -- & 56.44 & -- & -- & -- & 69.85 & -- & -- & -- & -- \\
\midrule
Baseline EN+FR & 1\text{B} & 61.44 & 15.36 & 55.79 & 57.83 & 52.32 & 50.77 & 66.86 & \textbf{71.24} & 57.93 & 58.36 & \textbf{62.29} & 62.24 \\
Baseline EN+FR & 360\text{M} & 61.44 & 15.36 & 55.26 & \underline{57.93} & 50.56 & 51.25 & 66.00 & 69.48 & 55.58 & 57.34 & 61.17 & 67.71 \\
\midrule
Cross-lingual Interleaving & 1\text{B} & 52.22 & 6.14 & 54.40 & 55.47 & 54.56 & 52.64 & 62.26 & 63.17 & \underline{63.28} & \underline{63.44} & 52.73 & 56.74 \\
Cross-lingual Interleaving & 360\text{M} & 52.22 & 6.14 & 55.90 & 57.08 & \textbf{56.44} & \textbf{55.37} & 64.00 & 68.67 & \textbf{65.20} & \textbf{65.84} & 55.35 & 59.56 \\
\midrule
Cross-Lingual Interleaving + EN+FR Fine-Tuning & 1\text{B} & 61.44 & 15.36 & 55.63 & \textbf{58.31} & \underline{55.21} & \underline{55.05} & \textbf{67.45} & 70.39 & 62.90 & 63.35 & 61.75 & 69.15 \\
Cross-Lingual Interleaving + EN+FR Fine-Tuning & 360\text{M} & 61.44 & 15.36 & 55.74 & 57.50 & 55.10 & 53.92 & 67.07 & \underline{70.55} & 59.86 & 62.28 & 61.08 & 68.62 \\
\bottomrule
\end{tabular}%
}
\end{table*}

\subsection{Cross\mbox{-}lingual interleaving helps monolingual capabilities}
\label{subsec:mono_results}

Starting from monolingual baselines in Table~\ref{tab:crosslingual_results}, French\mbox{-}only training already acquires non\mbox{-}trivial semantic competence with scores $56.44$\% and $69.85$\% on \emph{sSC} and \emph{sTC} for the $360$M model, respectively. The $1$B model shows the same level of competence. Interestingly, in this data regime the $360$M model often matches or exceeds the $1$B model on French, a pattern consistent with a capacity–data mismatch: with limited French hours, the smaller model is easier to optimise and less prone to overfitting or optimisation instabilities. Relative to English baselines, the \emph{sSC} task tends to be slightly more favourable to French, while \emph{sTC} is generally easier overall and, for French, yields an average increase of $2.64$ points over English.

Introducing interleaving \emph{without} a final stabilisation phase produces a predictable trade-off. Semantic scores in English decrease by a small margin (e.g., for the $1$B model, from $56.06$\% to $54.40$\%.), whereas syntactic and lexical probes (\emph{sBLiMP}/\emph{sWUGGY}) drop more noticeably (from $61.96$\% and $69.92$\% to $52.73$\% and $56.74$\%). An intuitive explanation is that exposure to rapidly alternating languages perturbs low-level regularities (agreement, morphology, phonotactics) more than high-level semantics. Despite this, French \emph{sSC} improves modestly even though the model sees substantially fewer French tokens than the monolingual baseline; in this regime, the $360$M model reaches $57.08$\% while the baseline with $3{\times}$ more tokens obtains $56.44$\%. The gains therefore cannot be attributed to data volume and are better explained as transfer from English into French. \emph{TopicCloze} in French may dip at this stage, suggesting that, on balance, raw interleaving trades a little monolingual polish for multilingual competence.

A brief bilingual stabilisation phase (fine\mbox{-}tuning on EN{+}FR without interleaving) restores most of the lost ground in English and consolidates the transfer into French. After stabilisation, English performance typically returns to within a hair’s breadth of the English\mbox{-}only baseline, particularly on lexical discrimination. Indeed, for the $1$B model, \emph{sBLiMP} and \emph{sWUGGY} recover to $61.75$\% and $69.15$\%, aligning with the EN baseline. At an equivalent French token budget, the stabilisation stage surpasses its monolingual FR baseline by a few points on both \emph{sSC} and \emph{sTC}, achieving $58.31$\% and $70.39$\% versus $55.31$\% and $67.07$\%. This two\mbox{-}stage recipe consistently benefits both model sizes.

\subsection{Cross\mbox{-}lingual abilities emerge with interleaving}
\label{subsec:xling_results}
Looking again at Table~\ref{tab:crosslingual_results} and the cross\mbox{-}lingual evaluations (EN$\to$FR and FR$\to$EN), mixed EN{+}FR training \emph{without} interleaving underperforms when prompted in one language and continued in the other. Focussing on the 360M results, the baseline achieves 50.56\% and 51.25\% on \emph{sSC} and 55.58\% and 57.34\% on \emph{sTC} (EN$\to$FR and FR$\to$EN, respectively). These scores are above chance, especially on \emph{sTC}. This gap likely reflects that \emph{sTC} is simpler than \emph{sSC}; thus, small but non\mbox{-}trivial gains on \emph{sSC} translate into larger gains on \emph{sTC}.

With interleaving, substantially larger cross\mbox{-}lingual performance is observed: \emph{sTC} approaches monolingual accuracy and \emph{sSC} narrows to within a few points. For the 360M interleaved model, \emph{sSC} reaches 56.44\% and 55.37\%, and \emph{sTC} reaches 65.20\% and 65.84\% (EN$\to$FR and FR$\to$EN). Comparing to the same model’s monolingual results, \emph{sSC} scores are 55.90\% (EN) and 57.08\% (FR), while \emph{sTC} scores are 64.00\% (EN) and 68.67\% (FR). These results show that (i) the model develops cross\mbox{-}lingual abilities that approach monolingual performance, and (ii) relative to the EN{+}FR baseline, interleaving yields markedly stronger cross\mbox{-}lingual continuation. The effect is particularly strong at 360M, indicating that smaller models, under limited French, extract proportionally more benefit from multilingual context. With the stabilisation phase, we preserve these cross\mbox{-}lingual abilities with a small degradation: 55.10\% and 53.92\% on \emph{sSC}, and 62.90\% and 63.35\% on \emph{sTC} (EN$\to$FR and FR$\to$EN), representing average relative degradations of 1.39 points (\emph{sSC}) and 4.45 points (\emph{sTC}), offset by gains in monolingual performance. The same trends hold for the 1B models.

Finally, we measured layer\mbox{-}wise cosine similarity between hidden states for 1{,}000 aligned EN–FR sentences and found that interleaving leads to consistently stronger cross\mbox{-}lingual alignment than training on mixed data without interleaving, with average cosine similarities of 0.73 (EN{+}FR baseline), 0.75 (interleaving), and 0.76 (interleaving{+}stabilisation) for the 1B models.

\section{Conclusions}
\label{sec:conclusion}

We introduce a new cross\mbox{-}lingual interleaving scheme that mixes speech tokens across sentence\mbox{-}aligned languages without textual supervision, and we release an EN\mbox{--}FR training dataset, \emph{TinyStories} (\(\sim\!42\)k), together with spoken \emph{StoryCloze} and \emph{TopicCloze} benchmarks for EN\mbox{--}FR semantic evaluation. Under a matched training\mbox{-}token budget on $360$M\mbox{--}$1$B SLMs, the method improves monolingual semantic accuracy, enables robust cross\mbox{-}lingual continuation, and strengthens cross\mbox{-}lingual hidden\mbox{-}state alignment. Because the method places minimal constraints on optimisation dynamics, it is well suited to scaling multilingual SLMs, thereby enabling models to understand and converse across languages.

\vfill\pagebreak
\bibliographystyle{IEEEbib}
\bibliography{Template}

@string{icassp = "Proc. ICASSP"}

@string{interspeech = "Proc. Interspeech"}

@string{neurips = "Proc. NeurIPS"}

@string{emnlp = "Proc. EMNLP"}

@string{iclr = "Proc. ICLR"}

@string{icml = "Proc. ICML"}

@string{acl = "Proc. ACL"}

@string{jmlr = "Proc. JMLR"}

@string{tmlr = "Proc. TMLR"}

@inproceedings{transformer,
    author = {Ashish Vaswani and
                  others},
    title = {{Attention is All You Need}},
    year = {2017},
    booktitle=neurips,
}

@misc{Radford2018ImprovingLU,
  title={{Improving Language Understanding by Generative Pre-Training}},
  author={Alec Radford and others},
  year={2018},
  url={https://cdn.openai.com/research-covers/language-unsupervised/language_understanding_paper.pdf}
}

@misc{radford2022robustspeechrecognitionlargescale,
      title={{Robust Speech Recognition via Large-Scale Weak Supervision}}, 
      author={Alec Radford and others},
      year={2022},
      eprint={2212.04356},
      archivePrefix={arXiv},
      primaryClass={eess.AS},
      url={https://arxiv.org/abs/2212.04356}, 
}

@inproceedings{zeng2024scalingspeechtextpretrainingsynthetic,
 author = {Zeng, Aohan and others},
 booktitle = iclr,
 pages = {49396--49419},
 title = {Scaling Speech-Text Pre-training with Synthetic Interleaved Data},
 url = {https://proceedings.iclr.cc/paper_files/paper/2025/file/7b5ae891000049b91b3b62de596b1560-Paper-Conference.pdf},
 volume = {2025},
 year = {2025}
}

@inproceedings{chou-etal-2023-toward,
    title = {{"Toward Joint Language Modeling for Speech Units and Text"}},
    author = "Chou, Ju-Chieh  and others",
    editor = "Bouamor, Houda  and
      Pino, Juan  and
      Bali, Kalika",
    booktitle = emnlp,
    month = dec,
    year = "2023",
    address = "Singapore",
    publisher = "Association for Computational Linguistics",
    url = "https://aclanthology.org/2023.findings-emnlp.438/",
    doi = "10.18653/v1/2023.findings-emnlp.438",
    pages = "6582--6593"
}

@article{nguyen2024spiritlm,
    title = "{S}pi{R}it-{LM}: Interleaved Spoken and Written Language Model",
    author = "Nguyen, Tu Anh  and others",
    journal = "Transactions of the Association for Computational Linguistics",
    volume = "13",
    year = "2025",
    address = "Cambridge, MA",
    publisher = "MIT Press",
    url = "https://aclanthology.org/2025.tacl-1.2/",
    doi = "10.1162/tacl_a_00728",
    pages = "30--52"
}

@article{pratap2024scaling,
  title={{Scaling speech technology to 1,000+ languages}},
  author={Pratap, Vineel and others},
  journal=jmlr,
  volume={25},
  number={97},
  pages={1--52},
  year={2024}
}

@inproceedings{futami2025scheduled,
  title     = {{Scheduled Interleaved Speech-Text Training for Speech-to-Speech Translation with LLMs}},
  author    = {Hayato Futami and others},
  year      = {2025},
  booktitle = interspeech,
  pages     = {36--40},
  doi       = {10.21437/Interspeech.2025-1595},
  issn      = {2958-1796},
}

@inproceedings{yoo-etal-2025-code-switching,
    title = "{Code-Switching Curriculum Learning for Multilingual Transfer in {LLM}s}",
    author = "Yoo, Haneul  and others",
    editor = "Che, Wanxiang  and others",
    booktitle = acl,
    month = jul,
    year = "2025",
    address = "Vienna, Austria",
    publisher = "Association for Computational Linguistics",
    url = "https://aclanthology.org/2025.findings-acl.407/",
    doi = "10.18653/v1/2025.findings-acl.407",
    pages = "7816--7836",
    ISBN = "979-8-89176-256-5"
}

@article{defossez2024moshispeechtextfoundationmodel,
      title={{Moshi: a speech-text foundation model for real-time dialogue}}, 
      author={Alexandre Défossez and others},
      year={2024},
      eprint={2410.00037},
      archivePrefix={arXiv},
      primaryClass={eess.AS},
      url={https://arxiv.org/abs/2410.00037}, 
      journal={arXiv preprint arXiv:2410.00037},
}

@article{chen2022wavlm,
  title={{WavLM: Large-scale self-supervised pre-training for full stack speech processing}},
  author={Chen, Sanyuan and others},
  journal={Proc. IEEE Journal of Selected Topics in Signal Processing},
  volume={16},
  number={6},
  pages={1505--1518},
  year={2022},
  publisher={IEEE}
}

@inproceedings{cuervo-marxer-2024-scaling,
    title = "{Scaling Properties of Speech Language Models}",
    author = "Cuervo, Santiago  and others",
    editor = "Al-Onaizan, Yaser  and
      Bansal, Mohit  and
      Chen, Yun-Nung",
    booktitle = acl,
    month = nov,
    year = "2024",
    address = "Miami, Florida, USA",
    publisher = "Association for Computational Linguistics",
    url = "https://aclanthology.org/2024.emnlp-main.21/",
    doi = "10.18653/v1/2024.emnlp-main.21",
    pages = "351--361"
}

@article{
mousavi2025discrete,
title={{Discrete Audio Tokens: More Than a Survey!}},
author={Pooneh Mousavi and others},
journal=tmlr,
issn={2835-8856},
year={2025},
url={https://openreview.net/forum?id=eqNchtvc6v},
note={}
}

@article{zeghidour2021soundstream,
  title={{SoundStream: An end-to-end neural audio codec}},
  author={Zeghidour, Neil and others},
  journal={Proc. IEEE/ACM Transactions on Audio, Speech, and Language Processing},
  volume={30},
  pages={495--507},
  year={2021},
  publisher={IEEE}
}

@article{lakhotia2021generative,
  title={{On Generative Spoken Language Modeling from Raw Audio}},
  author={Lakhotia, Kushal and others},
  journal=acl,
  volume={9},
  pages={1336--1354},
  year={2021},
  publisher={MIT Press One Rogers Street, Cambridge, MA 02142-1209, USA journals-info~…}
}

@inproceedings{kharitonov2021text,
    title = "Text-Free Prosody-Aware Generative Spoken Language Modeling",
    author = "Kharitonov, Eugene  and others",
    editor = "Muresan, Smaranda  and
      Nakov, Preslav  and
      Villavicencio, Aline",
    booktitle = acl,
    month = may,
    year = "2022",
    address = "Dublin, Ireland",
    publisher = "Association for Computational Linguistics",
    url = "https://aclanthology.org/2022.acl-long.593/",
    doi = "10.18653/v1/2022.acl-long.593",
    pages = "8666--8681"
}

@article{nguyen2023generative,
  title={Generative spoken dialogue language modeling},
  author={Nguyen, Tu Anh and others},
  journal=acl,
  volume={11},
  pages={250--266},
  year={2023},
  publisher={MIT Press One Broadway, 12th Floor, Cambridge, Massachusetts 02142, USA~…}
}

@inproceedings{
park2024long,
title={{Long-Form Speech Generation with Spoken Language Models}},
author={Se Jin Park and Julian Salazar and Aren Jansen and Keisuke Kinoshita and Yong Man Ro and RJ Skerry-Ryan},
booktitle=icml,
year={2025},
url={https://openreview.net/forum?id=4AmFA0qNQ2}
}

@article{hassid2023textually,
  title={{Textually Pretrained Speech Language Models}},
  author={Hassid, Michael and others},
  journal=neurips,
  volume={36},
  pages={63483--63501},
  year={2023}
}

@article{borsos2023audiolm,
  title={{AudioLM: a Language Modeling Approach to Audio Generation}},
  author={Borsos, Zal{\'a}n and others},
  journal={Proc. IEEE/ACM transactions on audio, speech, and language processing},
  volume={31},
  pages={2523--2533},
  year={2023},
  publisher={IEEE}
}

@article{cuervo2025textspeechlanguagemodelsimproved,
      title={Text-Speech Language Models with Improved Cross-Modal Transfer by Aligning Abstraction Levels}, 
      author={Santiago Cuervo and others},
      year={2025},
      eprint={2503.06211},
      archivePrefix={arXiv},
      primaryClass={cs.CL},
      url={https://arxiv.org/abs/2503.06211}, 
      journal={arXiv preprint arXiv:2503.06211},
}

@inproceedings{maimon2025slammingtrainingspeechlanguage,
    title = "Slamming: Training a Speech Language Model on One {GPU} in a Day",
    author = "Maimon, Gallil  and others",
    booktitle = acl,
    month = jul,
    year = "2025",
    address = "Vienna, Austria",
    publisher = "Association for Computational Linguistics",
    url = "https://aclanthology.org/2025.findings-acl.631/",
    doi = "10.18653/v1/2025.findings-acl.631",
    pages = "12201--12216",
    ISBN = "979-8-89176-256-5"
}

@inproceedings{lin2024align,
    title = "Align-{SLM}: Textless Spoken Language Models with Reinforcement Learning from {AI} Feedback",
    author = "Lin, Guan-Ting  and others",
    booktitle = acl,
    month = jul,
    year = "2025",
    address = "Vienna, Austria",
    publisher = "Association for Computational Linguistics",
    url = "https://aclanthology.org/2025.acl-long.997/",
    doi = "10.18653/v1/2025.acl-long.997",
    pages = "20395--20411",
    ISBN = "979-8-89176-251-0"
}

@article{zeng2024scaling,
  title={{Scaling speech-text pre-training with synthetic interleaved data}},
  author={Zeng, Aohan and others},
  journal={arXiv preprint arXiv:2411.17607},
  year={2024}
}

@article{
defossez2022high,
title={{High Fidelity Neural Audio Compression}},
author={Alexandre D{\'e}fossez and others},
journal=tmlr,
issn={2835-8856},
year={2023},
url={https://openreview.net/forum?id=ivCd8z8zR2},
note={Featured Certification, Reproducibility Certification}
}

@article{eldan2023tinystories,
  title={Tinystories: How small can language models be and still speak coherent english?},
  author={Eldan, Ronen and others},
  journal={arXiv preprint arXiv:2305.07759},
  year={2023}
}

@article{zeghidour2025streamingsequencetosequencelearningdelayed,
      title={Streaming Sequence-to-Sequence Learning with Delayed Streams Modeling}, 
      author={Neil Zeghidour and others},
      year={2025},
      eprint={2509.08753},
      archivePrefix={arXiv},
      primaryClass={cs.CL},
      url={https://arxiv.org/abs/2509.08753}, 
      journal={arXiv preprint arXiv:2509.08753},
}

@article{dubey2024llama,
  title={The llama 3 herd of models},
  author={Dubey, Abhimanyu and others},
  journal={arXiv e-prints},
  pages={arXiv--2407},
  year={2024}
}

@inproceedings{mostafazadeh-etal-2016-corpus,
    title = "A Corpus and Cloze Evaluation for Deeper Understanding of Commonsense Stories",
    author = "Mostafazadeh, Nasrin  and others",
    editor = "Knight, Kevin  and
      Nenkova, Ani  and
      Rambow, Owen",
    booktitle = acl,
    month = jun,
    year = "2016",
    address = "San Diego, California",
    publisher = "Association for Computational Linguistics",
    url = "https://aclanthology.org/N16-1098/",
    doi = "10.18653/v1/N16-1098",
    pages = "839--849"
}

@inproceedings{kang2024libriheavy,
  title={Libriheavy: A 50,000 hours ASR corpus with punctuation casing and context},
  author={Kang, Wei and others},
  booktitle=icassp,
  pages={10991--10995},
  year={2024},
  organization={IEEE}
}

@inproceedings{nguyen2020zero,
  title     = {The Zero Resource Speech Challenge 2021: Spoken Language Modelling},
  author    = {Ewan Dunbar and others},
  year      = {2021},
  booktitle = interspeech,
  pages     = {1574--1578},
  doi       = {10.21437/Interspeech.2021-1755},
  issn      = {2958-1796},
}

@article{team2024qwen2,
  title={Qwen2 technical report},
  author={Team, Qwen},
  journal={arXiv preprint arXiv:2407.10671},
  year={2024}
}

@article{brown2020language,
  title={Language models are few-shot learners},
  author={Brown, Tom and others},
  journal=neurips,
  volume={33},
  pages={1877--1901},
  year={2020}
}

@article{achiam2023gpt,
  title={{GPT-4} technical report},
  author={Achiam, Josh and others},
  journal={arXiv preprint arXiv:2303.08774},
  year={2023}
}

@article{kingma2014adam,
  title={Adam: A method for stochastic optimization},
  author={Kingma, Diederik P and Ba, Jimmy},
  journal={arXiv preprint arXiv:1412.6980},
  year={2014}
}

@article{gu2021efficiently,
  title={Efficiently modeling long sequences with structured state spaces},
  author={Gu, Albert and Goel, Karan and R{\'e}, Christopher},
  journal={arXiv preprint arXiv:2111.00396},
  year={2021}
}

\end{document}